\documentclass{article}

 \usepackage[final]{neurips_2025_ml4ps}

\usepackage[utf8]{inputenc} 
\usepackage[T1]{fontenc}    
\usepackage{hyperref}       
\usepackage{url}            
\usepackage{booktabs}       
\usepackage{amsfonts}       
\usepackage{nicefrac}       
\usepackage{microtype}      
\usepackage{xcolor}         
\usepackage{multirow}
\usepackage{graphicx}
\usepackage{subcaption}
\usepackage{amsmath}
\usepackage{pdflscape}
\usepackage{tikz}

\title{Enforcing governing equation constraints in neural PDE solvers via training-free projections}

\author{%
  Omer Rochman Sharabi \quad Gilles Louppe \\
  \\
  University of Liège
}

\newcommand{\uhat}{\hat{u}}
\newcommand{\JF}{J_h}

\begin{document}

\maketitle

\begin{abstract}

Neural PDE solvers used for scientific simulation often violate governing equation constraints. While linear constraints can be projected cheaply, many constraints are nonlinear, complicating projection onto the feasible set. Dynamical PDEs are especially difficult because constraints induce long-range dependencies in time. In this work, we evaluate two training-free, post hoc projections of approximate solutions: a nonlinear optimization-based projection, and a local linearization-based projection using Jacobian-vector and vector-Jacobian products. We analyze constraints across representative PDEs and find that both projections substantially reduce violations and improve accuracy over physics-informed baselines.

\end{abstract}

\section{Introduction}

Neural PDE solvers for scientific simulations often break physical constraints, even when they score well on standard metrics~\citep{zhengInverseBenchBenchmarkingPlugandPlay2025, chen2023Machine, yin2024Precipitation}. In particle systems this could mean violating mass and momentum conservation~\citep{Prantl2022GuaranteedCO}; in fluids, incompressibility and adherence to Navier–Stokes~\citep{boxho2025Turbulent}; in medical imaging, physiologically plausible ranges for parameters~\citep{ahmadi2025Physicsinformed}. Depending on the system, physical constraints can be linear or nonlinear, local or global, static or dynamical. 

In many problems, physical consistency is best understood as satisfying the constraints implied by the governing PDE, the problem specification, and known physics. Per-timestep conditions, such as divergence-free in incompressible fluids, restrict admissible states but can leave temporal consistency underdetermined, since they hold at each step independently. Conservation laws can help but do not guarantee consistent time evolution. Treating the governing equation itself as a constraint couples space and time and addresses this gap, providing a limit on temporal consistency. In this case, the PDE residual often serves as an informative and, sometimes, the only feasible metric of time consistency, because it dictates how plausible states evolve. Nonetheless, even local near-consistency can result in trajectories that are inconsistent globally.

Existing constraint enforcement methods involve significant tradeoffs. Physics-informed approaches like PINNs~\citep{raissi2017pinn}, PINO~\citep{li2023pino}, and their generative counterparts~\citep{zhou2025Generating} add penalty terms to the loss function, complicating training and potentially preventing convergence~\citep{sophiya2025comprehensive}. Architectural enforcement guarantees constraint satisfaction~\citep{mattheakis2019physical} but reduces model expressiveness and requires custom designs for each constraint type. Helper networks~\citep{constanteflores2025enforcing} introduce additional training overhead and hyperparameter tuning. 
Constrained sampling methods for generative models~\citep{huang2024DiffusionPDE, jacobsen2024CoCoGen, baldan2025Flow, luGenerativeDownscalingPDE2024} require backpropagation through sampling paths with repeated constraint evaluations. Moreover, these approaches are rarely validated on complex dynamical constraints like Navier–Stokes, focusing instead on simpler linear constraints such as divergence-free conditions. Finally, PDE determinism can lead to degenerate posteriors with unique solutions, making sampling from such narrow distributions challenging.

\textbf{Contributions.} We focus on hard governing equation constraints from nonlinear dynamical PDEs that induce long-range temporal dependencies. We evaluate two training-free, post hoc projection operators of approximate solutions: a nonlinear optimization-based projection and a local linearization-based projection implemented with Jacobian-vector and vector-Jacobian products for scalability.
In experiments, both projections substantially reduce violations and improve accuracy over physics-informed training baselines.

\section{Projection operators}

We consider a PDE of the general form
\begin{equation} \label{eq:pde}
\mathcal{L}[u]=f,
\end{equation}
over $\Omega\times(0,T)$, where $u$ is the unknown function we seek to find, $\Omega$ is the spatial domain, $(0, T)$ is the time interval, $\mathcal{L}$ is a differential operator that may be nonlinear, and $f$ is a known source function on $\Omega \times (0, T)$. The PDE is subject to initial and boundary conditions $\mathcal{I}[u]=u_0$ and $\mathcal{B}[u]=g$, where $\mathcal{I}$ and $\mathcal{B}$ are the initial and boundary condition operators, $u_0$ specifies the initial state on $\Omega$, and $g$ specifies the boundary values on $\partial \Omega \times (0, T)$.

To enforce global consistency, the full solution must be considered at once.  Let the discretized state $u \in \mathbb{R}^{N}$ be a vector of function evaluations at $N$ points in space and time,
$u = \{ u(x_i, t_i) \}_{i=1}^N$,
where the points $\{(x_i, t_i) \in \Omega \times (0, T)\}_{i=1}^N$ are chosen to be sufficiently dense to ensure accurate operator discretization. Note that we overload $u$ to denote both the continuous function and its discretized representation. The discretized PDE problem becomes a system of nonlinear equations
$h(u) = c \in \mathbb{R}^N$
where $h(u) = [\mathcal{L}_d(u), \mathcal{B}_d(u), \mathcal{I}_d(u)]$ is the concatenation of the discretized operators and $c = [f_d, g_d, u_{0,d}]$ is the concatenation of the discretized right-hand sides. 

Given the output $\uhat$ of a neural PDE solver,  we seek a physically consistent solution $u$ that satisfies the constraint $h(u) = c$. If consistency is defined in the $L_2$  sense, then this can be formulated as finding the projection of $\uhat$ into the feasible manifold,
\begin{equation}\label{eq:projection}
    \min_{u}\|u - \uhat\| \quad \text{s.t.} \quad h(u)=c.
\end{equation}

\textbf{Nonlinear projection.} When $h$ is nonlinear, there is no closed form solution to Equation~\ref{eq:projection}. Instead, the problem can be formulated as an unconstrained minimization problem
\begin{equation}\label{eq:min}
    \min_{u}\|u - \uhat\|+\lambda\|h(u)-c\|,
\end{equation}
with weight $\lambda \geq 0$, which can be solved by numerical optimization. We use the LBFGS algorithm~\citep{lbfgs} throughout this work.

\textbf{Linear projection.} As Equation~\ref{eq:projection} admits analytical solutions when $h$ is linear, an alternative is to linearize the constraints about $\uhat$ and rearrange
\begin{equation}\label{eq:system}
h(u)\approx h(\uhat)+ \JF (u-\uhat) \Rightarrow \JF u = c - h(\uhat) + \JF \uhat,
\end{equation}
where $\JF$ is the Jacobian of $h$ computed at $\uhat$. We denote this linear system as $\mathcal{C}u=b$, where $\mathcal{C} = \JF$ and $b =  c - h(\uhat) + \JF \uhat$. With linearized constraints, we formulate either the projection of $\uhat$ onto the feasible set (Equation~\ref{eq:exact}) or, if we suspect $\uhat$ to be a poor approximation, the relaxed variant (Equation~\ref{eq:relaxed})

\vspace{-1em}
\noindent\begin{minipage}[t]{0.48\linewidth}
\begin{equation}\label{eq:exact}
\min_{u}\|u - \uhat\|\quad\text{s.t.}\quad \mathcal{C}u=b
\end{equation}
\end{minipage}\hfill
\begin{minipage}[t]{0.48\linewidth}
\begin{equation}\label{eq:relaxed}
\min_{u}\|u - \uhat\|+\lambda\|\mathcal{C}u-b\|.
\end{equation}
\end{minipage}

Both admit closed-form solutions~\citep{nocedal}

\vspace{-1em}
\noindent\begin{minipage}[t]{0.48\linewidth}
\begin{equation}\label{eq:exact_sol}
u=\uhat-\mathcal{C}^{\top}(\mathcal{C}\mathcal{C}^{\top})^{-1}(\mathcal{C}\uhat-b)
\end{equation}
\end{minipage}\hfill
\begin{minipage}[t]{0.48\linewidth}
\begin{equation}\label{eq:relaxed_sol}
u=(I+\lambda\mathcal{C}^{\top}\mathcal{C})^{-1}(\uhat+\lambda\mathcal{C}^{\top}b).
\end{equation}
\end{minipage}

The first is a projection enforcing the linearized constraints exactly when they are consistent, while the second trades constraint satisfaction and proximity to $\uhat$ through $\lambda \geq 0$. Therefore, we call the first projection \emph{constrained} and the second \emph{relaxed}. Depending on the size of the system, $\mathcal{C}$ can be too large to fully materialize, and matrix inversion is known to be unstable~\citep{LA1, LA2, LA3}. In this case, instead of direct inversion, we solve the linear systems

\vspace{-1em}
\noindent\begin{minipage}[t]{0.48\linewidth}
\begin{equation}\label{eq:exact_system}
\mathcal{C}\mathcal{C}^{\top}X=\mathcal{C}\uhat-b,\qquad
\end{equation}
\end{minipage}\hfill
\begin{minipage}[t]{0.48\linewidth}
\begin{equation}\label{eq:relaxed_system}
(I+\lambda\mathcal{C}^{\top}\mathcal{C})X=\uhat+\lambda\mathcal{C}^{\top}b,
\end{equation}
\end{minipage}

and respectively compute $u = \uhat - \mathcal{C}^\top X$ or $u = X$, using sparse solvers such as CG~\citep{cg}, GMRES~\citep{gmres}, or BiCGSTAB~\citep{bicgstab}. These solvers only require access to the matrix-vector operators, so we leverage the fact that Equations~\ref{eq:exact_system} and~\ref{eq:relaxed_system} can be rewritten in terms of the $\text{JVP}(v):= v \mapsto \mathcal{C} v$ and $\text{VJP}(v):= v \mapsto \mathcal{C}^\top v$ to obtain favorable scaling. Alternatively, for moderately sized systems the sparse matrix $\mathcal{C}$ can be used directly.

\section{Experiments}

We investigate three dynamical systems: the 3D Lorenz ODE~\citep{lorenz63}, the 1D Kuramoto–Shivashinsky (KS) PDE~\citep{kuramoto, shivashinksy}, and the 2D Navier–Stokes (NS) PDE. More details on data generation are given in App.~\ref{app:data}. In the KS and NS PDEs, we train a neural network to predict the full trajectory \(x_{1:T}\) from the initial state \(x_0\), whereas in the Lorenz system we give $x_T$ as additional context to the network. For every equation we train two models: one that minimizes MSE only, and a physics-informed one that minimizes the MSE plus the MSE of the constraint loss $h(u) - c$. As physics-informed models, PINNs~\citep{raissi2017pinn} are used for Lorenz, while PINOs~\citep{li2023pino} are used for KS and NS. 
We report test-set Mean Squared Error (MSE) and the squared \(L_2\) norm of the constraint violation $\| h(u) - c \|$ (i.e. the residual) of the output of the networks without further processing, and after nonlinear and linear projections. For the NS equation, we show how the solution changes as the constraint violation reduces, and give a rough notion of the local constraint landscape. The results are shown in Table~\ref{tab:results}, with more details in App.~\ref{app:results}. 

\textbf{Lorenz.} An MLP and a PINN-MLP are trained on the Lorenz equation. For the relaxed projection, $\lambda = 1000$ was used. All projections cut constraint violation by $70\%$ or more and lower the MSE. The PINN-MLP has higher constraint violation despite having been trained to minimize it, as it did not train as effectively as the standard MLP under the same time and resource budget. 

\textbf{Kuramoto-Shivashinsky.} A Fourier Neural Operator (FNO)~\citep{li2021fourier} and a PINO were trained at a \(64\) resolution and evaluated at \(64\), \(128\), and \(256\) resolutions. Because KS dynamics are sufficiently resolved over these grids, the MSE is resolution-invariant, but constraint error rises with grid size. For the relaxed projection, $\lambda = 10$ was used. Projections suppress the violation at all resolutions and also reduce MSE. 

\begin{figure}
    \centering
    \includegraphics[width=0.75\linewidth]{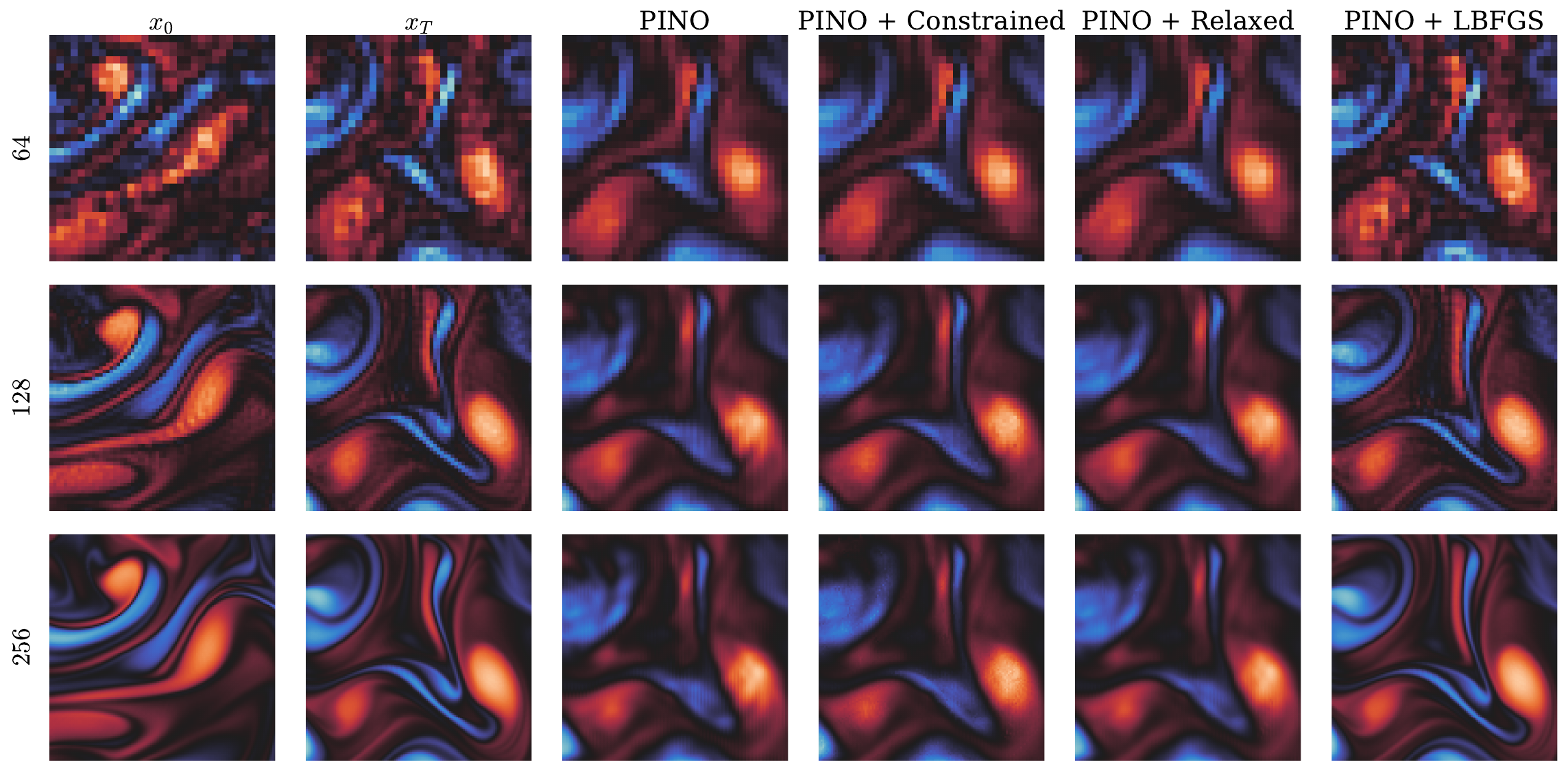}
    \vspace{-2pt}
    \caption{NS predictions comparing a PINO baseline with projection methods. Column 1 shows the initial condition $x_0$, Column 2 shows the ground truth $x_T$, and Columns 3-6 show predictions from PINO and projection methods. 
    Only LBFGS successfully recovers the fine-scale structures; physics-informed models fail to capture details they never encountered during training. }    
    \label{fig:ns}
\end{figure}

\begin{figure}
    \centering
    \includegraphics[width=0.65\linewidth]{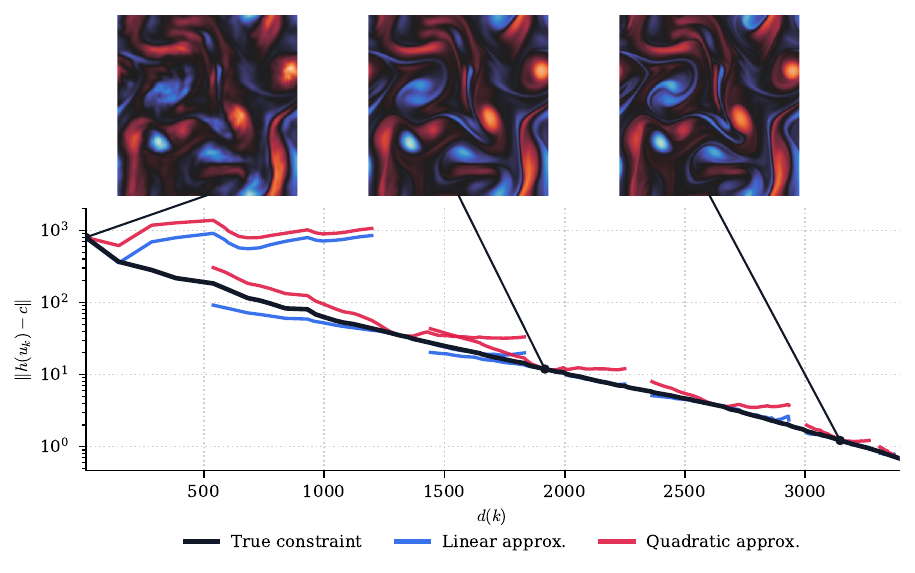}
    \vspace{-6pt}
    \caption{Constraint violation along LBFGS optimization path $u_{0:200}$ from initial guess $u_0 = \hat u$ to final solution $u^* = u_{200}$ (black, log scale), where the $x$-axis  measures the size $d(k) = \sum_{i=0}^k \|u_{i+1} - u_i \|$ of the path $u_{0:k}$, for $k=0$ to $200$.
    Blue and red curves show 1st and 2nd order Taylor approximations of the violation. The curves reveal two regimes: linear approximations work well initially but become inaccurate near the solution, while quadratic approximations remain reliable throughout, explaining why LBFGS outperforms linear projection methods.}    

    \label{fig:struct}
\end{figure}

\textbf{Navier-Stokes.} Similarly to KS, an FNO and a PINO were trained at a \(64\) resolution and evaluated at \(64\), \(128\), and \(256\). For the relaxed projection, $\lambda = ||\uhat||$ was used. Across resolutions, the baseline MSE stays nearly constant because large-scale structures captured by the model dominate the error. Unlike KS, NS contains smaller-scale details, which shows up as constraint violations and missing fine structure in the model output. Once those violations are reduced, the MSE drops significantly, as shown in Figure~\ref{fig:ns} and Table~\ref{tab:results}. In Figure~\ref{fig:struct} we show the local structure of the constraint by measuring its violation along the LBFGS trajectory, helping explain why LBFGS outperforms the two linearization-based projections, as seen in Table~\ref{tab:results}. They are good locally at the beginning but quickly degrade, and close to the minimum the linear approximation becomes noisy.

\begin{table*}[t]
\centering
\setlength{\tabcolsep}{6pt}
\resizebox{\textwidth}{!}{
\begin{tabular}{lcccccc}
\toprule
 & \multicolumn{2}{c}{\textbf{Lorenz}} & \multicolumn{2}{c}{\textbf{Kuramoto-Shivashinsky (64)}} & \multicolumn{2}{c}{\textbf{Navier--Stokes (64)}} \\
\cmidrule(lr){2-3}\cmidrule(lr){4-5}\cmidrule(lr){6-7}
 & \textbf{MSE} ($\times10^{-1}$)$\downarrow$ & \textbf{Residual} ($\times10^{-4}$)$\downarrow$
 & \textbf{MSE} ($\times10^{-2}$)$\downarrow$ & \textbf{Residual} ($\times10^{-5}$)$\downarrow$
 & \textbf{MSE} ($\times10^{-1}$)$\downarrow$ & \textbf{Residual} ($\times10^{-2}$)$\downarrow$ \\
\midrule
\multicolumn{7}{l}{\textbf{Data-driven (baseline = MLP for Lorenz, FNO for KS/NS)}} \\
Baseline & 7.78 & 50.8 & 4.01 & 46.8 & 13 & 8.13 \\
Baseline + Constrained & \textbf{7.75} & 13.1 & 3.92 & 16 & 10.5 & 4.21 \\
Baseline + Relaxed & 7.76 & 13.3 & 4 & 39.9 & 12.2 & 3.4 \\
Baseline + LBFGS & 7.76 & \textbf{1.18} & \textbf{3.91} & \textbf{4.53} & \textbf{2.63} & \textbf{0.00901} \\
\addlinespace
\multicolumn{7}{l}{\textbf{Physics-informed (PI = PINN for Lorenz, PINO for KS/NS)}} \\
PI & 9.6 & 51.7 & 4.76 & 4.27 & 13.8 & 6.48 \\
PI + Constrained & \textbf{9.57} & 13.4 & 4.77 & 24.9 & 11.5 & 3.75 \\
PI + Relaxed & 9.58 & 13.5 & 4.76 & 3.05 & 13 & 2.94 \\
PI + LBFGS & 9.58 & \textbf{1.23} & \textbf{4.75} & \textbf{1.27} & \textbf{3.21} & \textbf{0.00956} \\
\bottomrule
\end{tabular}}
\caption{Results of the experiments. The evolution for KS and NS is reported on resolution 64. Lower is better. LBFGS achieves significantly lower constraint violation. In NS, where the solutions have small scale details, the MSE improvement is much more significant.}
\label{tab:results}
\end{table*}

\section{Conclusion}

We demonstrated that training-free projection methods substantially outperform physics-informed training for enforcing governing equation constraints across multiple dynamical systems. LBFGS achieves near-perfect constraint satisfaction when the initial neural network output $\hat{u}$ provides a reasonable approximation, while linearization-based methods offer computational efficiency at the cost of accuracy. Our analysis reveals that linear approximations are locally effective but degrade rapidly away from the linearization point, suggesting that iterative relinearization approaches, similar to Sequential Quadratic Programming~\citep{nocedal}, could improve performance if implemented efficiently.

Several challenges remain for broader adoption. Current projection methods, while computationally tractable, cannot be directly integrated as differentiable layers within neural networks due to the computational cost of backpropagation through the optimization process. Reducing projection overhead is essential for scaling to larger systems and enabling end-to-end training. Additionally, extending these techniques to generative models presents opportunities for improving temporal coherence by projecting sampled trajectories onto physically consistent manifolds. Future work should focus on developing differentiable projection operators and exploring their integration with modern generative architectures.

\subsubsection*{Acknowledgments}
This work was partially supported by ServicePublic de Wallonie Recherche under grant n° 2010235 – ARIAC by DIGITALWALLONIA4.AI.

\bibliography{references}
\bibliographystyle{unsrt}

\appendix
\section{Appendix} \label{app:data}

\subsection{\textsc{Lorenz} data}

The Lorenz dataset is generated by integrating the 3-D Lorenz–63~\citep{lorenz63} system
\begin{equation}
\dot x = \sigma (y - x), \quad
\dot y = x (\rho - z) - y, \quad
\dot z = x y - \beta z .
\end{equation}
With $\sigma = 10$, $\rho = 28$, and $\beta = 8/3$. Trajectories are generated on-the-fly with \(x_0\sim\mathcal N(0,0.1)\). Euler integration was used with a fixed step $\mathrm dt = 1/512$ for $T = 512$ steps, giving a time horizon $t_{\max}=1$. Independent random seeds create disjoint training, validation, and test splits.

\subsection{\textsc{Kuramoto-Shivashinsky} data}

The 1-D Kuramoto–Shivashinsky equation~\citep{kuramoto, shivashinksy}
\begin{equation}
\partial_t u + \partial_{xx} u + \partial_{xxxx} u + u\,\partial_x u = 0
\end{equation}
is integrated on the periodic domain $x\in[0,64]$ with a pseudospectral solver. Spatial derivatives use the FFT on grids with $N\in\{64,128,256\}$ points and no de-aliasing. Time stepping was done using the first-order backward-difference formula (BDF1) with fixed step $\mathrm dt=10^{-1}$ for $512$ steps, giving $t_{\max}=51.2$. We generated \(65{,}536\) train, \(128\) validation, and \(128\) test simulations. During training, we used a subset of each solution, starting at step $256$ and ending at step $512$, corresponding to $t \in (25.6, 51.2)$. Each implicit step is solved by Newton iterations. Initial data are random sums of ten cosine modes,
\begin{equation}
u_0(x)=\sum_{k=1}^{10} a_k \cos\bigl(2 \pi\omega_k x/64+\phi_k\bigr),
\end{equation}
where amplitudes $a_k\sim\mathcal U(0,1)$, frequencies $\omega_k \sim \mathcal{U}(1, 5)$ and phases $\phi_k\sim\mathcal U(0,2\pi)$ are sampled independently. The dataset comprises $65{,}536$ training, $128$ validation, and $128$ test simulations (total $65{,}792$), each generated with an independent initial condition and solver seed.

\subsection{\textsc{Navier-Stokes} data}

The 2-D incompressible Navier--Stokes equations in vorticity form
\begin{equation}
\begin{aligned}
\frac{\partial \omega}{\partial t} &= - \mathbf{u}\cdot \nabla \omega + \frac{1}{\mathrm{Re}} \Delta \omega + f,\\
\mathbf{u} &= \left( \frac{\partial \psi}{\partial y}, - \frac{\partial \psi}{\partial x} \right),\\
\Delta \psi &= \omega,
\end{aligned}
\end{equation}
with the forcing  $f(\omega; x, y, t) = -4 \cos(4y) - 0.1 \omega(x,y,t)$ are integrated on the periodic domain $[0,2\pi]^2$ with $\mathrm{Re}=1000$ using a pseudospectral solver. Spatial derivatives use the FFT on grids with $N\in{64,128,256}$ and $2/3$ de-aliasing. Time stepping uses the explicit Heun method with fixed step $\mathrm dt = 2\times 10^{-3}$. We record $256$ snapshots at uniform intervals, giving a time horizon $T=8.192$. Initial data are zero-mean Gaussian random fields sampled at the highest resolution and projected to each grid. We generated $4096$ train, $128$ validation, and $128$ test simulations.

\section{Additional results} \label{app:results}

\begin{table}[h]
\centering
\small
\begin{tabular}{lcc}
\toprule
 & \textbf{MSE} ($\times10^{-1}$) & \textbf{Residual} ($\times10^{-4}$) \\
\midrule
MLP & 7.78 & 50.8 \\
MLP + Constrained & \textbf{7.75} & 13.1 \\
MLP + LBFGS & 7.76 & \textbf{1.18} \\
MLP + Relaxed & 7.76 & 13.3 \\
\addlinespace
MLP-PINN & 9.6 & 51.7 \\
MLP-PINN + Constrained & \textbf{9.57} & 13.4 \\
MLP-PINN + LBFGS & 9.58 & \textbf{1.23} \\
MLP-PINN + Relaxed & 9.58 & 13.5 \\
\bottomrule
\end{tabular}
\caption{Lorenz results. (Left) SE between GTs and generated trajectories, averaged over all trajectories and timesteps. (Right) Mean squared $L_2$ norm of the constraint violation of generated trajectories.}\label{app:tab:lor}
\end{table}

\begin{table}[h]
\centering
\small
\label{tab:ks}
\begin{tabular}{lcccccc}
\toprule
 & \multicolumn{3}{c}{\textbf{MSE} ($\times10^{-2}$)} & \multicolumn{3}{c}{\textbf{Residual} ($\times10^{-5}$)} \\
\cmidrule(lr){2-4}
\cmidrule(lr){5-7}
 & 64 & 128 & 256 & 64 & 128 & 256 \\
\midrule
FNO & 4.01 & 3.97 & 3.98 & 46.8 & 267 & 273 \\
FNO + Constrained & 3.92 & 3.72 & 3.73 & 16 & 71.4 & 74.6 \\
FNO + LBFGS & \textbf{3.91} & \textbf{3.65} & \textbf{3.66} & \textbf{4.53} & \textbf{3.73} & \textbf{3.77} \\
FNO + Relaxed & 4 & 3.87 & 3.88 & 39.9 & 182 & 185 \\
\addlinespace
PINO & 4.76 & 5.56 & 5.55 & 4.27 & 874 & 896 \\
PINO + Constrained & 4.77 & 5.42 & 5.01 & 24.9 & 368 & 239 \\
PINO + LBFGS & \textbf{4.75} & \textbf{4.67} & \textbf{4.66} & \textbf{1.27} & \textbf{1.11} & \textbf{1.18} \\
PINO + Relaxed & 4.76 & 5.18 & 5.17 & 3.05 & 502 & 509 \\
\bottomrule
\end{tabular}
\caption{Results for Kuramoto-Shivashinsky. (Left) SE between GTs and generated trajectories, averaged over all trajectories and timesteps. (Right) Mean squared $L_2$ norm of the constraint violation of generated trajectories. Constraint error increases with the resolution unless projection is performed.}\label{app:tab:ks}
\end{table}

\begin{table}[h]
\centering
\small
\label{tab:ns}
\begin{tabular}{lcccccc}
\toprule
 & \multicolumn{3}{c}{\textbf{MSE} ($\times10^{-1}$)} & \multicolumn{3}{c}{\textbf{Residual} ($\times10^{-2}$)} \\
\cmidrule(lr){2-4}
\cmidrule(lr){5-7}
 & 64 & 128 & 256 & 64 & 128 & 256 \\
\midrule
FNO & 13 & 9.09 & 8.85 & 8.13 & 6.58 & 6.53 \\
FNO + Constrained & 10.5 & 7.4 & 7.36 & 4.21 & 3.1 & 2.97 \\
FNO + LBFGS & \textbf{2.63} & \textbf{2.52} & \textbf{2.58} & \textbf{0.00901} & \textbf{0.00793} & \textbf{0.00813} \\
FNO + Relaxed & 12.2 & 8.41 & 8.18 & 3.4 & 2.9 & 2.88 \\
\addlinespace
PINO & 13.8 & 9.91 & 9.61 & 6.48 & 6.17 & 6.13 \\
PINO + Constrained & 11.5 & 8.27 & 8.13 & 3.75 & 3.07 & 2.97 \\
PINO + LBFGS & \textbf{3.21} & \textbf{3.01} & \textbf{3.02} & \textbf{0.00956} & \textbf{0.00829} & \textbf{0.00832} \\
PINO + Relaxed & 13 & 9.24 & 8.95 & 2.94 & 2.76 & 2.74 \\
\bottomrule
\end{tabular}
\caption{Results for Navier-Stokes. (Left) SE between GTs and generated trajectories, averaged over all trajectories and timesteps. (Right) Mean squared $L_2$ norm of the constraint violation of generated trajectories.}\label{app:tab:ns}
\end{table}

\end{document}